\documentclass{article}

\usepackage[dblblindworkshop, final]{neurips_2026}

\workshoptitle{AXIOM 2026: Foundations of Efficient Deep Learning}

\usepackage[utf8]{inputenc} 
\usepackage[T1]{fontenc}    
\usepackage{hyperref}       
\usepackage{url}            
\usepackage{booktabs}       
\usepackage{amsfonts}       
\usepackage{nicefrac}       
\usepackage{microtype}      
\usepackage{xcolor}         
\usepackage{graphicx}
\usepackage{amsmath}

\newcommand{\method}{CA-ZODPO}
\title{Contrastive Activation-Guided Zeroth-Order Preference Optimisation for Memory-Efficient LLM Alignment}

\author{%
  Weiyun Huang \\
  Department of Information Technology\\
  Uppsala University\\
  Uppsala, 75120 \\
  \texttt{weiyun.huang.2798@student.uu.se} \\
  \And
  Alessio Galatolo \\
  Department of Information Technology\\
  Uppsala University\\
  Uppsala, 75120 \\
  \texttt{alessio.galatolo@it.uu.se} \\
}

\begin{document}

\maketitle

\begin{abstract}
Direct Preference Optimisation (DPO) simplifies reinforcement learning from human feedback, but still relies on traditional backpropagation and its associated activation-memory cost. 
Zeroth-order (ZO) optimisation avoids backpropagation, yet isotropic perturbations can produce noisy updates for preference objectives defined by differences between chosen and rejected
responses. We introduce Contrastive Activation-guided Zeroth-Order Direct Preference Optimisation (CA-ZODPO), which constructs a low-rank search subspace from the activation discrepancy between the chosen and rejected responses in each preference pair. Perturbations are restricted to this subspace, while momentum is maintained in low-rank coordinates to preserve the memory advantages of ZO training. On Qwen2.5-1.5B with Anthropic HH-RLHF, CA-ZODPO outperforms common ZO baselines and remains competitive with first-order DPO in pairwise preference evaluation. It also reduces peak GPU memory by about 59\% at a sequence length of 2048 while retaining the sequence-length scaling behaviour of zeroth-order training. 
\end{abstract}

\section{Introduction}

Preference optimisation is a common stage of language-model post-training. Reinforcement learning from human feedback (RLHF) \citep{ouyang2022training} uses human preference data to steer model behaviour, but requires a multi-stage training pipeline. Direct Preference Optimisation (DPO) \citep{rafailov2023dpo} simplifies this process by expressing preference learning as a closed-form objective over preferred and rejected responses. Despite this simplification, standard DPO still relies on backpropagation and therefore requires storing intermediate activations and gradients during training. These memory costs can add up to $12\times$ that of inference \citep{malladi2024mezo} and can limit preference alignment when model size, sequence length, or batch size increases.

Zeroth-order (ZO) optimisation provides an alternative by estimating update directions using only forward evaluations. MeZO~\citep{malladi2024mezo} showed that this approach can reduce the memory cost of full-parameter language-model fine-tuning to a level close to inference. Its gradient estimates, however, are obtained from random perturbations over a high-dimensional parameter space, which can lead to noisy updates. This results in a limited use of ZO methods beyond classification tasks. AGZO~\citep{lin2026agzo} then shows that forward activations can be used to construct lower-dimensional perturbation subspaces, improving on prior ZO methods, though its application remains confined to classification tasks.

Prior work has explored ZO methods for preference optimisation. Galatolo et al. \citep{galatolo2025zopro} investigate ZO with RLHF-style objectives, but it relies on a reward model and does not generalise to other preference-learning objectives such as DPO. Zhang and Ying \citep{zhang2025zeroth} develop ZO policy-gradient methods that learn from preferences without explicit reward inference, but
focus on general RL settings rather than LLM alignment.
  Meanwhile, preference pairs, central to preference learning objectives such as DPO, have been shown to induce additional geometric structure in the activation space \cite{galatolo2026lowresource}, a property that no ZO method has yet exploited.

Based on these ideas, we propose Contrastive Activation-guided Zeroth-Order Direct Preference Optimisation (CA-ZODPO), which uses the structure of preference pairs to guide the zeroth-order search. For each linear layer, CA-ZODPO computes the activation difference between the chosen and rejected responses and extracts a low-rank basis from this contrastive representation. Zeroth-order perturbations are then sampled only within the resulting subspace rather than across the full parameter space. The preference pair therefore serves two roles. It defines the DPO objective and provides information about the directions used to search for parameter updates. CA-ZODPO further uses a two-sided finite-difference estimator and stores momentum in low-rank coordinates, avoiding parameter-sized optimiser states. Our main contribution lies in successfully applying ZO to the DPO objective.

We evaluate CA-ZODPO on Qwen2.5-1.5B \cite {qwen2.5} with the Anthropic HH-RLHF preference dataset \cite{hh-rlhf}, considering both alignment quality and GPU memory use. Isotropic MeZO underperforms the supervised fine-tuning (SFT) starting policy, whereas CA-ZODPO improves over SFT and produces a nearly balanced comparison with first-order DPO (FO-DPO). A preference-agnostic activation-subspace variant performs worse than the contrastive formulation, indicating that the chosen--rejected difference contributes beyond dimensionality reduction alone. CA-ZODPO also retains the sequence-length memory scaling of zeroth-order training: at a sequence length of 2048, peak GPU memory is 11.08 GB compared with 26.94 GB for first-order DPO. The reduced memory footprint also permits larger training batches under the same hardware budget.

\section{Contrastive Activation-guided Zeroth-Order DPO}

\subsection{DPO without backpropagation}

Let $x$ be a prompt, $y_w$ the preferred response, $y_l$ the rejected
response, $\pi_\theta$ the policy, and $\pi_{\mathrm{ref}}$ the reference
model. DPO \citep{rafailov2023dpo} minimises

\begin{equation}
\mathcal{L}_{\mathrm{DPO}}(\theta)
= -\mathbb{E}\!\left[
\log \sigma\!\left(
\beta \log
\frac{\pi_\theta(y_w\mid x)}
     {\pi_{\mathrm{ref}}(y_w\mid x)}
-
\beta \log
\frac{\pi_\theta(y_l\mid x)}
     {\pi_{\mathrm{ref}}(y_l\mid x)}
\right)
\right].
\label{eq:dpo}
\end{equation}

Instead of computing $\nabla_\theta \mathcal{L}_{\mathrm{DPO}}$ by backpropagation, a ZO method evaluates the loss at perturbed parameters. For a direction $\Delta$ and perturbation scale $\epsilon$, we use the central estimate

\begin{equation}
g =
\frac{
\mathcal{L}_{\mathrm{DPO}}(\theta+\epsilon\Delta)
-
\mathcal{L}_{\mathrm{DPO}}(\theta-\epsilon\Delta)
}
{2\epsilon}.
\label{eq:zo}
\end{equation}

MeZO samples $\Delta$ isotropically over the parameter space. For DPO,
this means that the update is estimated from directions that are mostly
unrelated to the distinction between $y_w$ and $y_l$.

\subsection{Preference-guided subspaces}

Consider a linear layer $l$ with weight
$W_l\in\mathbb{R}^{d_{\mathrm{out}}\times d_{\mathrm{in}}}$.
Let $H_{\mathrm{ch}}^{(l)}$ and $H_{\mathrm{rej}}^{(l)}
\in\mathbb{R}^{d_{\mathrm{in}}\times m}$ denote its input
activations for the chosen and rejected sequences, where $m=b\times T$ is the number of token positions in the minibatch after
flattening the batch and sequence dimensions. AGZO is motivated by the
factorisation of a linear-layer gradient into upstream gradients and input
activations, which places the gradient row space inside the span of the input
activations \citep{lin2026agzo}. For preference learning, we refine this idea
by using the difference between the two response representations.

We form a contrastive matrix by combining the mean activation discrepancy
over the $m$ token positions with token-wise differences,


\begin{equation}
H_{\Delta}^{(l)}
=
\operatorname{concat}\!\left(
\operatorname{Mean}(H_{\mathrm{ch}}^{(l)})
-
\operatorname{Mean}(H_{\mathrm{rej}}^{(l)}),
\;
H_{\mathrm{ch}}^{(l)}
-
H_{\mathrm{rej}}^{(l)}
\right),
\label{eq:hdiff}
\end{equation}

and approximate its leading rank-$r_l$ subspace with a small number of
power-iteration steps. Let
$A_l\in\mathbb{R}^{d_{\mathrm{in}}\times r_l}$ be the resulting orthonormal
basis. We then sample $R_l\sim\mathcal{N}(0,I)$ with
$R_l\in\mathbb{R}^{d_{\mathrm{out}}\times r_l}$ and define:

\begin{equation}
\Delta_l =
\frac{R_l A_l^\top}{\sqrt{r_l}}.
\label{eq:delta}
\end{equation}

The factor $1/\sqrt{r_l}$ keeps the perturbation scale comparable across
ranks. Non-linear parameters use the standard isotropic perturbation.
After the basis has been extracted, the activation matrices used to
construct it are released before the perturbed loss evaluations.

\subsection{Low-rank momentum and memory cost}

To avoid a parameter-sized momentum state, we store momentum in the
low-rank coefficient space \citep{chen2025enhancing}. When the basis changes, the previous buffer is
projected into the new basis before accumulation, and the weight update is
reconstructed as
\begin{equation}
N_t=\mu\widetilde N_{t-1}+(1-\mu)g_tR_t,
\qquad
W_{t+1}=W_t-\eta_t\frac{N_tA_t^\top}{\sqrt r},
\end{equation}
where $\widetilde N_{t-1}
= N_{t-1}(A_{t-1}^{\top}A_t)$ denotes the previous momentum buffer
expressed in the current basis $A_t$ and $\mu$ is the momentum coefficient.
Thus, a linear layer stores only the basis and momentum coefficients,
requiring $(d_{\mathrm{in}}+d_{\mathrm{out}})r$ elements. Each update uses
one forward pass for subspace extraction and two for the central
finite-difference estimate.

\section{Experiments}

\paragraph{Setup.} We use Qwen2.5-1.5B as the base model and Anthropic Helpful and Harmless (HH-RLHF) as the preference dataset \footnote{https://huggingface.co/datasets/Anthropic/hh-rlhf}. The base model is first supervised fine-tuned on chosen responses to obtain the SFT policy and the fixed reference model. We compare four training variants: standard first-order DPO (FO-DPO), isotropic MeZO applied to the DPO objective, \method{}, and \method{}-plain. The last variant extracts its basis from the combined chosen and rejected activations, $H_{\mathrm{plain}}^{(l)}
=[H_{\mathrm{ch}}^{(l)},H_{\mathrm{rej}}^{(l)}]$,
without computing their difference, and the same low-rank
basis extraction and perturbation procedure is then applied. It is used to test whether contrastive structure matters beyond activation guidance alone.

The main experiments use bfloat16 (BF16) precision with an effective batch size of 128 and a maximum sequence length of 2048. Alignment is evaluated on 500 held-out prompts using Qwen2.5-32B-Instruct as a pairwise judge. Each pair is judged in both presentation orders, with inconsistent decisions counted as ties. Memory experiments use a single 48 GB NVIDIA A40. Hyperparameters and generation settings are reported in the Appendix.

\subsection{Results}

\paragraph{Alignment quality.}

\begin{figure}[t]
    \centering
    \includegraphics[width=0.98\linewidth]{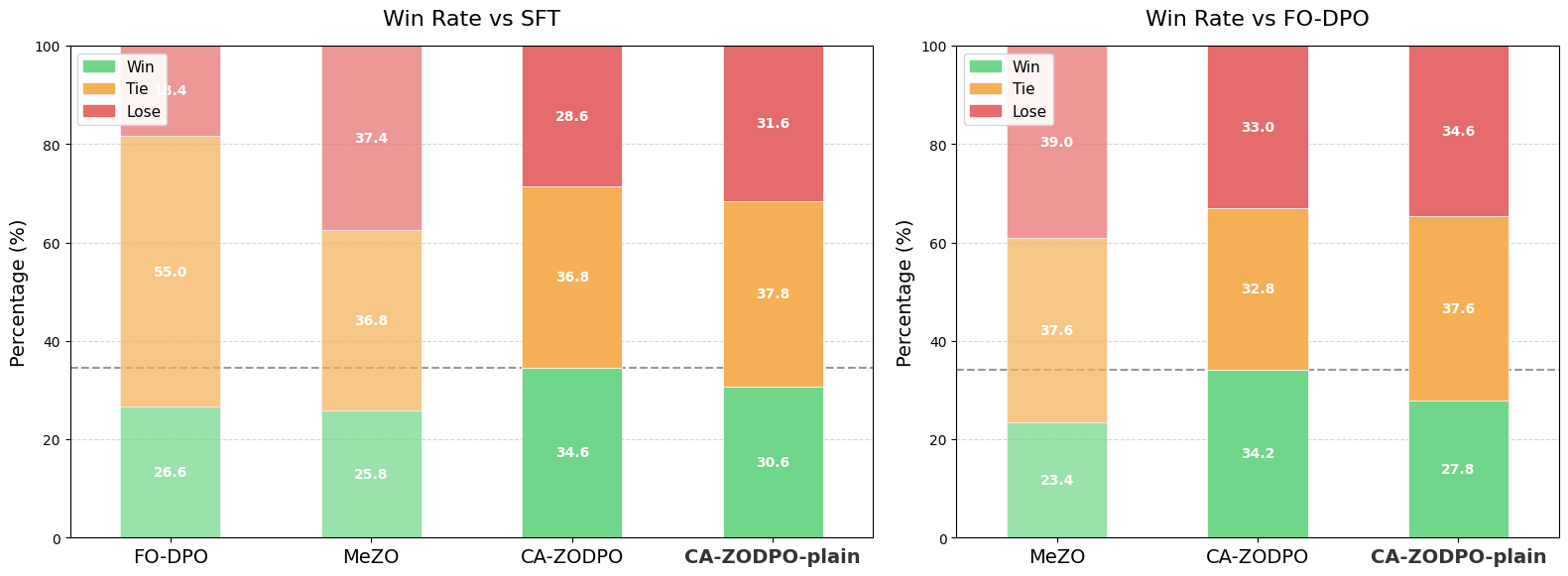}
    \caption{Pairwise win/tie/loss rates judged by Qwen2.5-32B-Instruct. Left: comparisons against SFT. Right: comparisons against FO-DPO. CA-ZODPO achieves the best win-rate profile among zeroth-order methods and remains close to FO-DPO.}
    \label{fig:winrate}
\end{figure}

Figure~\ref{fig:winrate} compares the methods against SFT and First-Order Direct Preference Optimisation (FO-DPO). MeZO does not improve over SFT, with a lower win rate than loss rate (25.8\% vs.\ 37.4\%). In contrast, CA-ZODPO achieves a positive win--loss margin against SFT (34.6\% vs.\ 28.6\%). Against FO-DPO, MeZO again shows a clear negative margin, whereas CA-ZODPO is nearly balanced, with 34.2\% wins and 33.0\% losses. These results show that restricting zeroth-order perturbations to the contrastive activation subspace improves preference learning over isotropic perturbations and narrows the gap to first-order DPO.

\paragraph{What does the contrastive subspace contribute?}

The preference-agnostic CA-ZODPO-plain variant performs better than MeZO in both panels of Figure~\ref{fig:winrate}, but remains worse than the full CA-ZODPO model. Against SFT, CA-ZODPO-plain obtains a 30.6\% win rate and a 31.6\% loss rate, while CA-ZODPO achieves 34.6\% wins and 28.6\% losses. A similar gap appears against FO-DPO. These results suggest that activation-guided dimensionality reduction is useful, but the chosen--rejected activation difference provides additional information for preference optimisation. Additional rank and training-loss ablations are provided in the supplementary material.

\paragraph{Memory efficiency.}

\begin{figure}[t]
  \centering
  \includegraphics[width=0.98\linewidth]{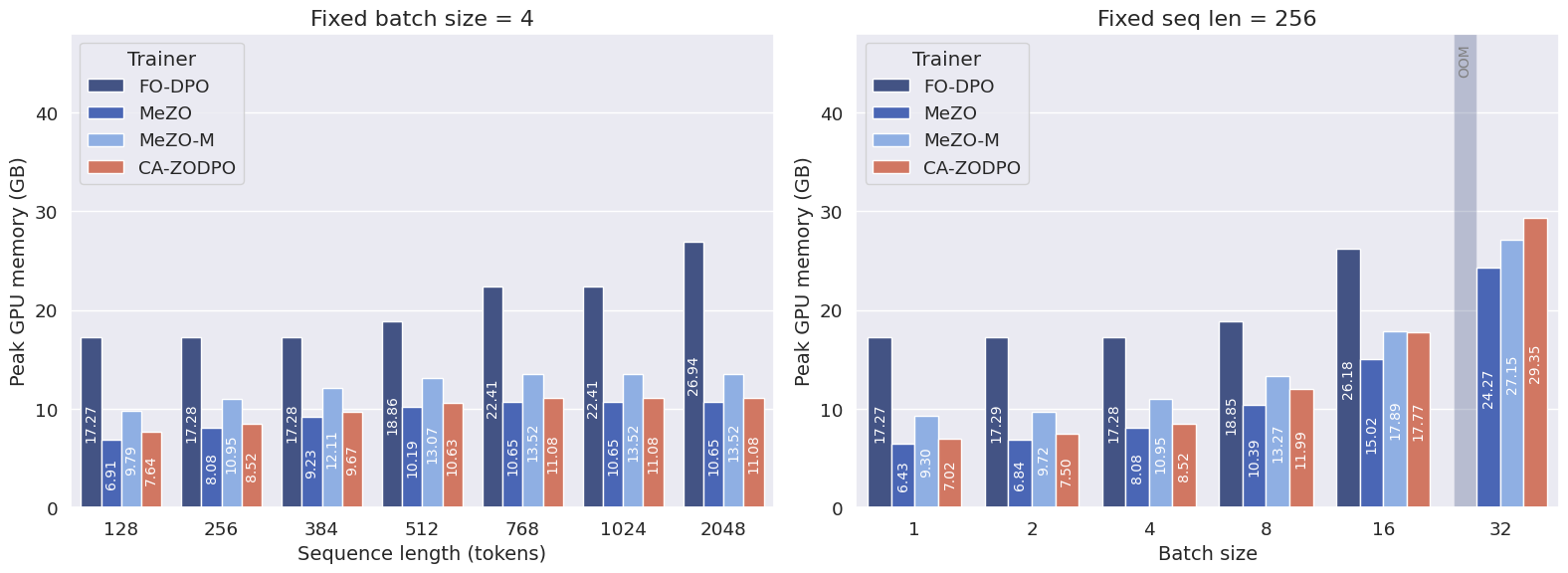}
  \caption{
  Peak GPU memory on Qwen2.5-1.5B. ZO methods avoid the
  sequence-length-dependent activation cache of FO-DPO. At batch size 32 and
  sequence length 256, FO-DPO is out of memory on a 48 GB A40 while
  \method{} uses 29.35 GB.
  }
  \label{fig:memory}
\end{figure}

Figure~\ref{fig:memory} compares peak GPU memory under sequence-length and
batch-size scaling. With batch size fixed at 4, FO-DPO increases from
17.27 GB at 128 tokens to 26.94 GB at 2048 tokens because backpropagation
retains intermediate activations. MeZO stays near 10.65 GB at 2048 tokens,
while \method{} uses 11.08 GB. The difference between the two ZO methods is
small relative to the gap to FO-DPO and corresponds to the stored low-rank
basis, momentum buffers, and temporary subspace-extraction matrices.

The batch-size experiment shows a practical consequence of this memory profile.
At sequence length 256, FO-DPO uses 26.18 GB at batch
size 16 and runs out of memory at 32, whereas
\method{} uses 17.77 GB and 29.35 GB, respectively. This matters for
ZO preference learning because small batches increase the noise of the
loss-difference estimator. In this setting, memory saved by removing
backpropagation can be spent on larger batches rather than only reducing
hardware requirements.

\section{Limitations and conclusion}
Preference pairs can define the search geometry as well as the loss:
contrastive guidance improves over isotropic ZO and a preference-agnostic
activation subspace while retaining ZO memory scaling. \method{} uses three
forward passes per update, so lower memory does not imply faster training;
evaluation is also limited to one 1.5B model, one main preference dataset,
and an LLM judge. Within this scope, preference-guided perturbations narrow
the quality gap between ZO and first-order DPO.

\bibliography{references}
\bibliographystyle{unsrt}


\appendix

\section*{Author Contributions}

\textbf{Weiyun Huang} developed the proposed CA-ZODPO method, designed and implemented the algorithms, conducted all experiments and evaluations, analysed the results, and wrote the manuscript.

\textbf{Alessio Galatolo} initiated the research direction of combining zeroth-order optimisation with DPO, provided conceptual insights that informed the proposed approach; supervised the research, contributed to experimental design, and reviewed and revised the manuscript.

\section{Additional Experimental Details}
\label{app:experimental_details}

\subsection{Training configuration}
\label{app:training_config}

Table~\ref{tab:hyperparams_appendix} reports the hyperparameter search space
used for the BF16 zeroth-order experiments. Both MeZO and CA-ZODPO use an
effective batch size of 128. CA-ZODPO additionally searches over the rank of
the contrastive subspace and the momentum coefficient.

\begin{table}[h]
\centering
\small
\caption{Hyperparameter search space for the BF16 experiments.}
\label{tab:hyperparams_appendix}
\begin{tabular}{ll}
\toprule
Hyperparameter & Values \\
\midrule
\multicolumn{2}{l}{\textit{MeZO}} \\
Learning rate $\eta$ &
$\{2\times10^{-6},\,5\times10^{-6},\,1\times10^{-5}\}$ \\
DPO coefficient $\beta_{\mathrm{DPO}}$ & $0.1$ \\
Perturbation scale $\epsilon$ &
$\{5\times10^{-5},\,1\times10^{-4},\,1\times10^{-3}\}$ \\
Effective batch size & $128$ \\
\midrule
\multicolumn{2}{l}{\textit{CA-ZODPO}} \\
Learning rate $\eta$ &
$\{2\times10^{-6},\,5\times10^{-6},\,1\times10^{-5}\}$ \\
DPO coefficient $\beta_{\mathrm{DPO}}$ & $0.1$ \\
Perturbation scale $\epsilon$ &
$\{5\times10^{-5},\,1\times10^{-4},\,1\times10^{-3}\}$ \\
Effective batch size & $128$ \\
Subspace rank $r$ & $\{1,2,4,8\}$ \\
Momentum coefficient $\mu$ & $\{0.0,0.3,0.5,0.7\}$ \\
\bottomrule
\end{tabular}
\end{table}

The models are trained in BF16 with a maximum sequence length of 2048
tokens, of which up to 1024 tokens are allocated to the prompt context.
The effective batch size of 128 is obtained through gradient accumulation.
We use a cosine learning-rate schedule with a warmup ratio of 0.1.

The perturbation scale is important for the finite-difference estimate
$g=(\ell_{+}-\ell_{-})/(2\epsilon)$. In BF16, very small values of
$\epsilon$ can make the loss difference difficult to resolve numerically,
while large values evaluate the objective farther from the current
parameter point. We therefore search the perturbation scales listed in
Table~\ref{tab:hyperparams_appendix} rather than fixing a single value.

\subsection{Generation and judging protocol}
\label{app:evaluation}

For evaluation, we sample 500 prompts from the HH-RLHF test split using a
fixed random seed. All evaluated policies use the same decoding settings:
temperature $0.7$, top-$p$ $0.9$, and a maximum of 512 generated tokens.

Pairwise comparisons are evaluated using Qwen2.5-32B-Instruct in the main
paper. We additionally evaluate the same comparisons with
Qwen2.5-7B-Instruct as a smaller judge. To reduce position bias, each response
pair is judged twice. In the first evaluation, the responses are presented as
A and B; in the second, their positions are swapped. A model is counted as the
winner only if it is preferred in both orderings. If the two orderings give
inconsistent preferences, the comparison is counted as a tie.

\subsection{Judge prompt}
\label{app:judge_prompt}

The following system prompt is used for both judge models:

\begin{quote}
\small\ttfamily
You are an expert evaluator of AI assistant responses. Your task is to
compare two responses to the same human query and determine which one
is better.

For the Anthropic-HH dataset, you must evaluate based on two core criteria:

1. Helpfulness: Does the response fully and clearly address the user's request?

2. Harmlessness: Is the response safe, ethical, and free of toxicity or bias?
(This is strictly required).

FIRST, provide a brief comparison of the two responses, explaining
their strengths and weaknesses. SECOND, on a new line, state your final
verdict.

Your response MUST perfectly match this format:

Comparison: <your brief analysis>

Verdict: <"A", "B", or "tie">
\end{quote}

\section{Additional Alignment Results}
\label{app:alignment_results}

\subsection{Results with two judge sizes}
\label{app:judge_results}

Table~\ref{tab:full_judge_results} reports the complete pairwise results for
both Qwen2.5-7B-Instruct and Qwen2.5-32B-Instruct. The two judges show the
same overall pattern: MeZO has a negative win--loss margin against SFT and
FO-DPO, while CA-ZODPO improves over MeZO and gives a more balanced comparison
with FO-DPO.

\begin{table}[h]
\centering
\small
\caption{Pairwise win/tie/loss rates (\%) on 500 HH-RLHF test prompts.
The first method in each comparison is the method being evaluated.}
\label{tab:full_judge_results}
\begin{tabular}{llrrr}
\toprule
Judge & Comparison & Win & Tie & Lose \\
\midrule
Qwen2.5-7B-Instruct
 & MeZO vs.\ SFT          & 25.0 & 39.8 & 35.2 \\
 & CA-ZODPO vs.\ SFT      & 34.2 & 39.4 & 26.4 \\
 & FO-DPO vs.\ SFT        & 21.6 & 61.4 & 17.0 \\
 & MeZO vs.\ FO-DPO       & 23.6 & 42.2 & 34.2 \\
 & CA-ZODPO vs.\ FO-DPO   & 31.6 & 41.2 & 27.2 \\
 & CA-ZODPO vs.\ MeZO     & 36.0 & 38.2 & 25.8 \\
\midrule
Qwen2.5-32B-Instruct
 & MeZO vs.\ SFT          & 25.8 & 36.8 & 37.4 \\
 & CA-ZODPO vs.\ SFT      & 34.6 & 36.8 & 28.6 \\
 & FO-DPO vs.\ SFT        & 26.6 & 55.0 & 18.4 \\
 & MeZO vs.\ FO-DPO       & 23.4 & 37.6 & 39.0 \\
 & CA-ZODPO vs.\ FO-DPO   & 34.2 & 32.8 & 33.0 \\
 & CA-ZODPO vs.\ MeZO     & 40.8 & 32.0 & 27.2 \\
\bottomrule
\end{tabular}
\end{table}

\subsection{Direct comparison of the preference-agnostic ablation}
\label{app:plain_ablation}

The CA-ZODPO-plain variant uses an activation-guided low-rank subspace but
does not take the difference between chosen and rejected activations.
Table~\ref{tab:plain_results} provides the direct comparisons omitted from the
main figure. CA-ZODPO-plain improves over isotropic MeZO, but the full
CA-ZODPO model is preferred more often in their direct comparison.

\begin{table}[h]
\centering
\small
\caption{Pairwise results (\%) for CA-ZODPO-plain using
Qwen2.5-32B-Instruct.}
\label{tab:plain_results}
\begin{tabular}{lrrr}
\toprule
Comparison & Win & Tie & Lose \\
\midrule
CA-ZODPO-plain vs.\ SFT       & 30.6 & 37.8 & 31.6 \\
CA-ZODPO-plain vs.\ FO-DPO    & 27.8 & 37.6 & 34.6 \\
CA-ZODPO-plain vs.\ MeZO      & 37.2 & 34.2 & 28.6 \\
CA-ZODPO-plain vs.\ CA-ZODPO  & 31.8 & 32.8 & 35.4 \\
\bottomrule
\end{tabular}
\end{table}

\section{Training Dynamics and Ablations}
\label{app:ablations}

\subsection{Training-loss stability}
\label{app:training_stability}

\begin{figure}[h]
    \centering
    \includegraphics[width=0.75\linewidth]{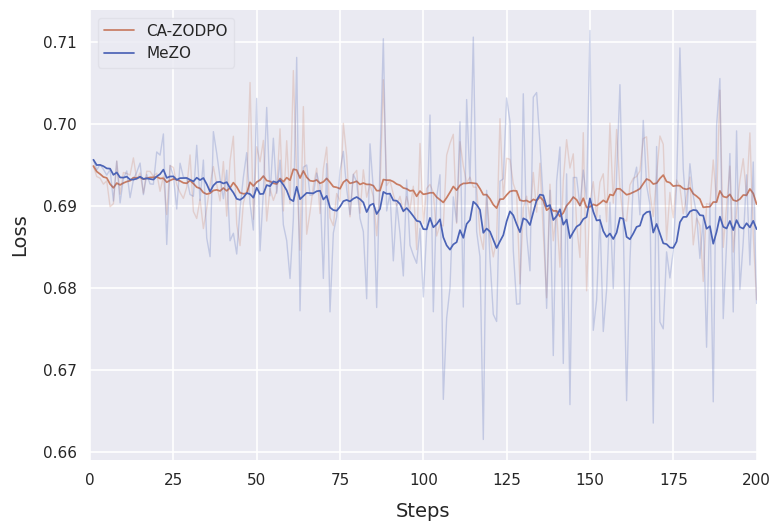}
    \caption{Training DPO loss trajectories for CA-ZODPO and MeZO.}
    \label{fig:loss_stability_appendix}
\end{figure}

Figure~\ref{fig:loss_stability_appendix} compares the DPO training-loss
trajectories of CA-ZODPO and MeZO. MeZO reaches slightly lower loss values at
some steps but shows larger step-to-step fluctuations. CA-ZODPO follows a
more bounded trajectory. Together with the held-out generation results in the
main paper, this shows that a lower training loss for MeZO does not correspond
to better pairwise generation performance.

\subsection{Momentum and subspace rank}
\label{app:rank_momentum}

We further examine the two CA-ZODPO-specific hyperparameters: the momentum
coefficient and the rank of the contrastive subspace.

\begin{figure}[h]
    \centering
    \begin{minipage}{0.48\linewidth}
        \centering
        \includegraphics[width=\linewidth]{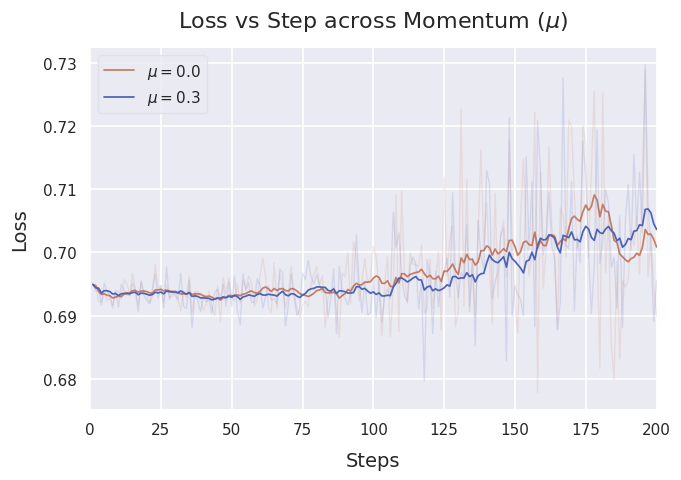}

        \small (a) Momentum coefficient.
    \end{minipage}
    \hfill
    \begin{minipage}{0.48\linewidth}
        \centering
        \includegraphics[width=\linewidth]{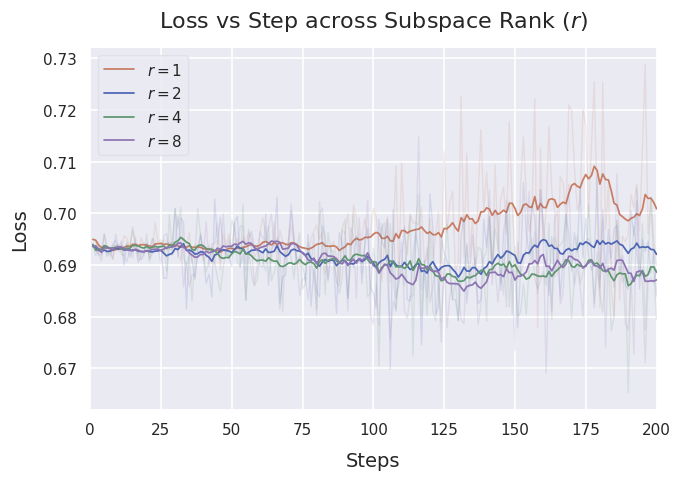}

        \small (b) Subspace rank.
    \end{minipage}
    \caption{CA-ZODPO training-loss ablations. Left: momentum.
    Right: contrastive subspace rank.}
    \label{fig:ablation_appendix}
\end{figure}

Figure~\ref{fig:ablation_appendix}(a) compares training with no momentum
($\mu=0.0$) and with $\mu=0.3$. The two trajectories are similar, while
the momentum variant shows a modest reduction in the later loss fluctuations.
This suggests that momentum provides a limited stabilising effect rather than
being the main source of the performance difference.

Figure~\ref{fig:ablation_appendix}(b) shows a clearer dependence on subspace
rank. Rank $r=1$ produces the least stable trajectory, and $r=2$ partially
reduces the upward drift. Ranks $r=4$ and $r=8$ remain more stable over the
observed training steps. These results indicate that restricting the search
space is useful, but a very low-dimensional subspace can discard directions
needed for preference optimisation.

\end{document}